\documentclass[10pt,twocolumn,letterpaper]{article}
\PassOptionsToPackage{table,dvipsnames}{xcolor}
\usepackage[T1]{fontenc}
\usepackage[pagenumbers]{paperstyle}
\usepackage{amsmath,amssymb,booktabs,multirow,array,tabularx,graphicx}
\newcolumntype{R}{>{\raggedleft\arraybackslash}X}
\usepackage{xcolor}
\definecolor{linkblue}{rgb}{0.16,0.38,0.58}
\definecolor{mpwash}{HTML}{FDF0E6}
\definecolor{tablewash}{HTML}{F5F5F3}
\definecolor{tablesoft}{HTML}{666660}
\usepackage[pagebackref,breaklinks,colorlinks,allcolors=linkblue]{hyperref}
\graphicspath{{figures/}}
\newcommand{\method}{MinPair}
\newcommand{\sg}{\operatorname{sg}}
\newcommand{\hit}{\operatorname{Hit}}
\title{\method{}: Measured Minimal Pairs for Fine-Grained Control\\in Text-to-Motion Generation}
\author{Lennart Schiweck \qquad Crist\'obal Curio\\
Cognitive Systems Group, Reutlingen University, Germany}
\begin{document}
\twocolumn[{%
\renewcommand\twocolumn[1][]{#1}%
\maketitle
\begin{minipage}{\textwidth}
\centering
\includegraphics[width=\textwidth]{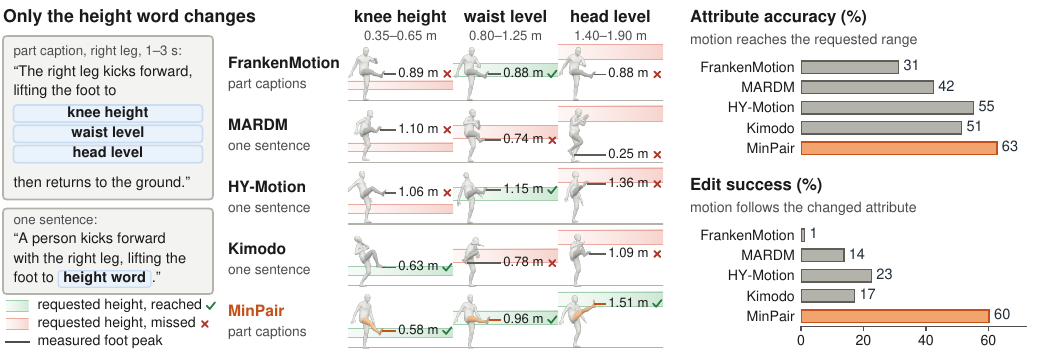}
\captionof{figure}{\textbf{Kick height requests.} Only the height word changes. Like FrankenMotion~\cite{li2026b}, \method{} reads the body-part caption of the right leg, whereas MARDM~\cite{meng2024mardm}, HY-Motion~\cite{wen2025hymotion} and Kimodo~\cite{rempe2026kimodo} read one sentence. Shown per model and height is the motion closest to the median foot peak of all motions generated from that prompt, with the requested range in green if the foot peak lies inside it and in red if it misses. The range below each height is the benchmark's fixed target range. MARDM, HY-Motion and Kimodo are SMPL-H~\cite{romero2017} fits to their stored joints. Only \method{} reaches all three ranges. Right, attribute accuracy over all 22 target tasks and the share of template edits followed, each with the input of its row (\cref{sec:setup}).}
\label{fig:teaser}
\end{minipage}
\vspace{6pt}

}]
\begin{figure*}[t]
\centering
\includegraphics[width=\textwidth]{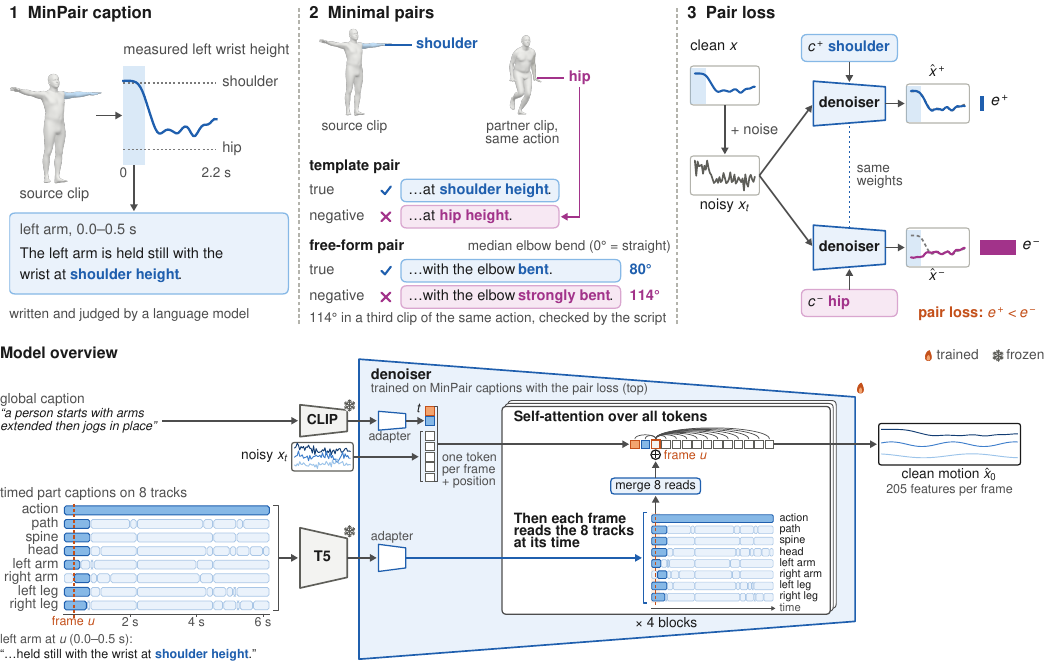}
\caption{\textbf{Method overview.} Top, training. (1)~A script measures quantities in a real clip, here the left wrist height. A language model writes them into a timed part caption and a further call judges it against the measurements. (2)~A template pair copies the true caption with the value of a partner clip of the same action, here hip height. A free-form pair rewords a short span after a measured difference such as the elbow bend and is checked against the measurements. (3)~Both captions condition the same denoiser on the same noisy motion. The pair loss asks the true caption for the smaller error inside the caption's time window (shaded). The curves in (3) are illustrative. Bottom, the part-track denoiser of \cref{sec:diffusion}, shown for frame $u$. Only the text encoders are frozen.}
\label{fig:overview}
\end{figure*}

\begin{abstract}
Text-to-motion generation has made great progress in capturing what action to perform, yet it remains limited in controlling how that action is executed. For example, when only the height word of a kick request changes, none of the tested state-of-the-art methods reaches all three of knee, waist and head height. The other words of a caption usually imply an attribute such as height, so neither the training loss nor the standard metrics require a model to use the word that states it. A model has to use this word only when it alone separates two otherwise identical captions. We introduce \method{}, which creates such caption pairs from real motion clips. It starts from rule-based motion descriptors that measure the 3D body motion of each clip. From these descriptors a language model writes \method{} captions, timed body-part captions that say how each body part moves. The same descriptors show in which attributes two clips of the same action differ. A measured minimal pair joins the true caption of one clip with an altered copy in which only one attribute takes the value measured in the other clip. A contrastive pair loss then requires a diffusion denoiser to reconstruct the clip more accurately from its true caption than from the altered copy. Training needs no synthetic motion and inference stays unchanged. Control is tested on 22 target tasks and on caption edits that each change one attribute. \method{} captions alone help motions reach their targets and improve the standard metrics M2M, R@3 and FID, yet a model trained on them still ignores most single-word edits. The pair loss raises the share of followed template edits from 11\% to 60\%, while the standard metrics do not improve further. With body-part captions, \method{} reaches more targets than state-of-the-art methods that were trained on up to 99 times more motion data.
\end{abstract}

\section{Introduction}
\label{sec:intro}
In text-to-motion generation~\cite{guo2022}, a request such as \emph{kick to head level} should produce a kick that reaches the head. The action alone leaves open how high, how fast and how far a movement goes and how much it turns. We call such execution properties \emph{attributes}. An animator who asks for a head-level kick or a robot operator who asks for \emph{three steps forward} expects exactly that execution. Diffusion models have become the standard generator for this task~\cite{ho2020ddpm,tevet2023}. FrankenMotion~\cite{li2026b} gives every body part its own timed track of sentences, which we call \emph{part captions}. HY-Motion~\cite{wen2025hymotion} and Kimodo~\cite{rempe2026kimodo} scale data and parameters. Yet when only the height word of a kick request changes, none of these state-of-the-art methods reaches all three requested heights (\cref{fig:teaser}). Nothing in their training singles out the attribute word. Text labels~\cite{punnakkal2021,guo2022} rarely state how high, how fast or how far a movement goes, so part captions derived from them~\cite{li2026b} rarely do either. Where captions do state it, the other words give it away. In captions written from measurements, a linear classifier recovers a masked tempo word from the remaining words of a held-out caption in 77\% of cases against 41\% for the most frequent tempo. The denoising loss and the standard metrics M2M, R@3 and FID all reward the complete caption, so a model can succeed from the context without using the attribute word. A model learns to follow a word reliably only once that word alone decides the outcome.

A motion corpus already records many executions of each action, such as kicks to knee and to head height. Measuring them shows which words fit each execution~\cite{delmas2022,liu2024a,yazdian2025motionscript} and which clips of the same action differ in a given attribute. Measured minimal pairs create these cases from real clips.

We introduce \method{} (\cref{fig:overview}). It measures every training clip with rule-based motion descriptors. From these a language model writes \emph{\method{} captions}, which say how each body part moves in each phase. For a source clip whose caption places the wrist at shoulder height, a partner clip of the same action keeps it at hip height. We copy the source caption and change only this value to the partner's. The copy is wrong for the source clip in exactly one attribute and serves as its \emph{negative caption}~\cite{yuksekgonul2023}. The new words arise in one of two ways. In a \emph{template pair}, the attribute word moves along a fixed word scale of the captions. In a \emph{free-form pair}, a language model words the measured difference itself in one span of up to 4 words, which we check against the measurements. Because the partner is a real clip, no motion is synthesized and the negative describes an execution that really occurs. As in contrastive learning~\cite{oord2018}, a \emph{pair loss} then requires the denoiser to reconstruct the same noisy copy of the source motion more accurately from the true caption than from the negative~\cite{krojer2023hardneg,lee2025softrepa}. Only the changed words can make this difference. Inference stays unchanged.

We test control with two benchmarks (\cref{sec:setup}). \emph{Target tasks} request explicit values such as a kick to waist level or a walk of three steps. \emph{Caption edits} change one attribute in the \method{} captions of held-out clips, as template or free-form edits built like the two kinds of pairs. With \method{} captions but without the pair loss, our denoiser follows only 11\% of the template edits. The pair loss raises this share to 60\% while M2M, R@3 and FID do not improve further. With body-part captions \method{} reaches more targets than state-of-the-art methods~\cite{meng2024mardm,wen2025hymotion,rempe2026kimodo,li2026b}, although HY-Motion and Kimodo were trained on 23 to 99 times more motion data (\cref{sec:compare}).

Our contributions are as follows.
\begin{itemize}
\item We introduce measured minimal pairs and a contrastive pair loss that make a single attribute of a caption controllable in text-to-motion diffusion models. Each pair combines the true caption of a clip with a negative caption that takes one value from another real clip of the same action, showing a targeted form of supervision that can outweigh scale (\cref{sec:pairs,sec:loss,sec:compare}).
\item We introduce \method{} captions, timed part captions written from measurements of each motion. They state the attributes of each body part in each phase. With the same network they improve attribute accuracy and all three standard metrics over previous part captions (\cref{sec:base,sec:results}).
\item We introduce two control benchmarks, target tasks and caption edits, which isolate single attribute words. They show that controlling a single attribute is a separate skill from fitting a complete caption, which M2M, R@3 and FID do not measure (\cref{sec:setup,sec:results}).
\end{itemize}

\section{Related Work}
\label{sec:related}
\paragraph{Text-to-motion generation and fine-grained text.}
HumanML3D~\cite{guo2022} became the standard benchmark for text-to-motion generation and MDM~\cite{tevet2023} made denoising diffusion the common generator. Recent work refines the text side. FineMoGen~\cite{zhang2023finemogen} generates motion from fine-grained spatio-temporal descriptions, STMC~\cite{petrovich2024stmc} composes timed prompts on a multi-track timeline during sampling and FrankenMotion~\cite{li2026b} conditions each body part on its own timed sentences. FineMotion~\cite{wu2025finemotion} and KinMo~\cite{zhang2025kinmo} describe the movements of single body parts or kinematic groups. FineXtrol~\cite{shen2026finextrol} and MATE~\cite{han2026mate} learn finer text representations with hierarchical or word-level contrastive objectives. Other work scales the generator. HY-Motion~1.0~\cite{wen2025hymotion} trains a billion-parameter flow-matching transformer on more than 3,000\,h of motion, Kimodo~\cite{rempe2026kimodo} trains a diffusion model on a large optical motion-capture corpus, OpenT2M~\cite{cao2026opent2m} scales data with detailed temporal annotation and MotionMaster~\cite{jiang2026motionmaster} unifies generation and editing. PoseScript~\cite{delmas2022}, PoseFix~\cite{delmas2023}, Kinematic Phrases~\cite{liu2024a} and MotionScript~\cite{yazdian2025motionscript} connect measured pose and motion properties to language and CoMo~\cite{huang2024como} generates and edits motion through pose codes of single body parts. LaMoGen~\cite{kim2025lamogen} guides a pretrained diffusion model with the effort and shape qualities of Laban Movement Analysis~\cite{laban1971}. \method{} builds on measured part captions and adds a training signal that makes the generator separate attribute values within one action.

\paragraph{Hard negatives and post-training.}
ARO~\cite{yuksekgonul2023} uses composition-aware negative captions to expose weaknesses of vision-language models and SugarCrepe~\cite{hsieh2023} shows that carelessly built negatives contain linguistic shortcuts. Diffusion Classifier~\cite{li2023diffusionclassifier} classifies an image by the text that gives the lowest denoising error. HardNeg-DiffusionITM~\cite{krojer2023hardneg} and SoftREPA~\cite{lee2025softrepa} train with contrastive objectives over denoising errors and TDSM~\cite{do2025tdsm} recognizes skeleton actions with a triplet loss over them. Fujiwara~\etal~\cite{fujiwara2024car} train motion-language models with negative texts whose events are shuffled in time. \method{} transfers the denoising-based principle to motion generation and mines its negatives from measured differences between two real executions of one action. MoDiPO~\cite{pappa2026modipo} adapts diffusion preference optimization to motion and MotionRFT~\cite{tan2026motionrft} applies reinforcement fine-tuning with learned motion rewards. OmniME~\cite{shi2026omnime} learns motion editing from positive and negative examples. \method{} needs neither a reward model nor generated candidates. It contrasts two captions of one real motion.

\paragraph{Evaluating instruction following.}
SATO~\cite{chen2024sato} tests stability under semantically similar texts. T2MBench~\cite{yang2026t2mbench} evaluates instruction following, physical plausibility and fine control across many prompts and models, with numeric targets for the whole body and for pairs of body parts. MotionFix~\cite{athanasiou2024motionfix} benchmarks motion editing with triplets of source motion, target motion and edit text. Our benchmarks add target ranges for single body parts and paired caption edits under identical sampling noise, which isolate the effect of one changed attribute.

\section{Method}
\label{sec:method}
\method{} measures every training motion, writes the measurements into timed part captions and pairs these captions with negative captions that each change one measured attribute (\cref{fig:overview}).

\subsection{Part-track diffusion model}
\label{sec:diffusion}
As in FrankenMotion~\cite{li2026b}, a caption consists of a global caption for the whole clip and eight tracks for action, path (the root trajectory), spine, head, left arm, right arm, left leg and right leg. We call one sentence with its time span a \emph{segment}. Tracks need not share their boundaries.

A motion clip $x\in\mathbb{R}^{T\times D}$ has $T$ frames with $D=205$ features each, in the motion representation of FrankenMotion. Its text condition $c$ is such a part caption. \method{} is a diffusion model~\cite{ho2020ddpm,tevet2023}. The forward process mixes $x$ with Gaussian noise,
\begin{equation}
 x_t=\sqrt{\bar\alpha_t}\,x+\sqrt{1-\bar\alpha_t}\,\epsilon,
 \qquad \epsilon\sim\mathcal{N}(0,I),
 \label{eq:forward}
\end{equation}
for a diffusion step $t\in\{1,\dots,100\}$ with a cosine schedule $\bar\alpha_t$. We call $x$ the clean motion and $x_t$ the noisy motion. The denoiser $f_\theta(x_t,t,c)$ predicts the clean motion from the noisy one. For a set $W$ of frame indices, such as the time span of one segment, its reconstruction error under caption $c$ is
\begin{equation}
 e_W(c)=\frac{1}{|W|}\sum_{u\in W}
 \bigl\|f_\theta(x_t,t,c)_u-x_u\bigr\|^2_{\omega},
 \label{eq:error}
\end{equation}
where $\omega$ weights the motion features as in FrankenMotion. The error covers all features of the frames in $W$, not only the joints of one body part. The denoising loss $\mathcal{L}_{\mathrm{denoise}}=\mathbb{E}_{(x,c),\,t,\,\epsilon}[e_M(c)]$ averages this error over the set $M$ of all valid frames of a clip with $t$ drawn uniformly.

\paragraph{Part-track denoiser.}
The denoiser is a Transformer over frames (\cref{fig:overview}, bottom). Every frame of $x_t$ becomes one token. Two more tokens carry the diffusion step $t$ and the global caption, which frozen CLIP~\cite{radford2021clip} encodes. Frozen T5-Large~\cite{raffel2020t5} encodes every segment. Trained adapters map both text features into the denoiser. Each of four blocks first applies self-attention over all tokens followed by a feed-forward layer. Then every frame reads the segment that is active at this frame on each of the eight tracks through cross-attention. A small MLP fuses the eight reads with flags that mark the active tracks and adds the result to the frame token. A linear head predicts the clean motion.

\subsection{\method{} captions}
\label{sec:base}
\method{} writes the seven body-part tracks of almost every caption from the motion itself. The global caption and the action track remain those of FrankenMotion~\cite{li2026b}, which a language model merged from existing text labels, mostly from BABEL~\cite{punnakkal2021} and HumanML3D~\cite{guo2022}.

First, we compute the joint positions of every clip with the body model and shape of its recording. Rule-based motion descriptors summarize them in a short text that states every 0.75 to 1\,s which limbs move, how high hands and feet rise, how the joints bend and how far the pelvis travels and turns. Qwen3.6-27B~\cite{qwen2026qwen36} reads this summary with a table of how much each body region moves and cuts the clip into phases of the whole body. It sees no text labels in this step. For every phase and body part we then turn the measured values into level words on fixed word scales, such as \emph{knee height}, \emph{shoulder height} or \emph{overhead} for heights and \emph{still}, \emph{slow}, \emph{moderate} or \emph{fast} for tempo. The distance travelled becomes a number of steps and the turn an angle rounded to 5$^\circ$.

A second call of the language model writes the sentences from the summary, the phases and these level words. The prompt asks for one sentence per body part and time span and for tracks without gaps. It asks to describe a repeated movement such as walking in one segment that states its count, to name the side of every arm and leg and to give heights and directions relative to the body, as in \emph{across the body}. It allows numbers only for turn angles and counts. The call also sees the BABEL and HumanML3D labels of the clip as hypotheses. It may name the activity after them only where the measurements support them for that body part and time span. A third call judges every sentence against the summary. Its verdict is stored for inspection and removes no sentence. The caption in \cref{fig:overview}~(1) arises this way from the wrist-height curve.

\method{} captions comprise 450,179 segments on 12,726 training clips, a median of 29 per clip. The same pipeline writes \method{} test captions for 1,583 test clips, which no model sees in training.

\subsection{Measured minimal pairs}
\label{sec:pairs}
A minimal pair consists of two captions of the same clip that differ only in the words of one attribute. The \emph{negative caption} copies the true caption of the source clip and changes only these words to the value of a partner clip, another real execution of the same action. All other words, segments and tracks stay the same. The partner only supplies this value and is never reconstructed. Template pairs and free-form pairs differ in how the new words arise.

\paragraph{Template pairs.}
In a template pair, the attribute word moves along a fixed word scale of the captions. We search for two training clips of the same action whose captions state different values of one attribute on the same track, such as the highest height word of a leg track, a tempo word, a turn angle or a step count. A value counts only when the descriptors of its clip confirm it. The partner's word then replaces the source word, for example \emph{hip height} for \emph{shoulder height} (\cref{fig:overview}~(2)). A second search compares two clips of the same action directly on 51 measured quantities. There a difference must exceed the measurement noise and a visibility threshold, such as half a height level of the person's body or 30$^\circ$ of turn. Template pairs cover height, tempo, turn, count and side and the four posture attributes trunk lean, trunk twist, head speed and knee bend.

\paragraph{Free-form pairs.}
In a free-form pair, the language model words the measured difference itself in one span of up to 4 words. This also covers differences that the word scales of the captions do not name, such as elbow bend or stance width. For each training clip we pick up to 2 partners of the same action and compare both clips on 61 measured quantities. The language model receives the source caption with its time spans and, for every visible difference, the descriptor values of both clips with their words. It returns a true and a negative sentence for each pair, where the negative changes one span to the partner's value. The true sentence may add the source value to a segment that does not state it and then also replaces this segment in the denoising loss. We check that the sentence lies on a track and time span where the difference occurs and that the changed words point from the source to the partner value wherever the check can read both values from the words. A rejected pair returns once to the language model with the reason. Of the proposed pairs 72.4\% pass.

Training uses 190,141 negative captions for 11,824 of the 12,822 training clips, 48,155 of them from free-form pairs.

\subsection{Pair loss}
\label{sec:loss}
For a negative caption $c^-_j$ with affected time window $W_j$, we compare the reconstruction errors of the same noisy motion under both captions,
\begin{equation}
 z_j=\frac{e_{W_j}(c^+)-e_{W_j}(c^-_j)}
 {\tau\max\{\sg[e_{W_j}(c^+)],10^{-8}\}},
 \label{eq:logit}
\end{equation}
where $c^+$ is the true caption or its paraphrase, $\sg$ stops the gradient and $\tau=0.1$. For a clip with $J$ negative captions the pair loss is
\begin{equation}
 \mathcal{L}_{\mathrm{pair}}=\log\Bigl(1+\sum_{j=1}^{J}\exp z_j\Bigr).
 \label{eq:pair}
\end{equation}
Dividing by the detached positive error makes the difference relative, so clips and diffusion steps with large errors do not dominate the loss.

\paragraph{Relation to contrastive learning.}
Define the score of a caption as $s_W(c)=-e_W(c)/(\tau\,\sg[e_W(c^+)])$. Then $z_j=s_{W_j}(c^-_j)-s_{W_j}(c^+)$. For negatives that share one window $W$, \cref{eq:pair} is exactly the InfoNCE loss~\cite{oord2018} with these scores over one positive and $J$ negative captions. Training is thus contrastive learning over captions in which the denoiser is the scoring function, as in HardNeg-DiffusionITM~\cite{krojer2023hardneg} and SoftREPA~\cite{lee2025softrepa} for images. \method{} mines its negatives from measurements and restricts each comparison to the window of the changed segment.

The objective adds the pair loss to the denoising loss,
\begin{equation}
 \mathcal{L}=\mathcal{L}_{\mathrm{denoise}}+
 \lambda\,\mathbb{E}_{\mathrm{sel}}\bigl[\mathcal{L}_{\mathrm{pair}}\bigr],
 \quad\lambda=0.05,
 \label{eq:total}
\end{equation}
where the pair loss is averaged over the selected clips of a batch and the denoising loss covers all clips.

\subsection{Training}
\label{sec:training}
We train the denoiser from random initialization with \cref{eq:total} for 1,500 epochs (300,000 steps at batch size 64). We use AdamW with a constant learning rate of $10^{-4}$. An exponential moving average of the weights (decay 0.9995) gives the evaluated model. A clip with negative captions is always eligible for the pair loss at the noisier steps $t\geq50$ and with probability 0.2 at the less noisy steps below. Each batch selects at most 21 of its 64 clips, those with negatives from a partner first. For template pairs, a paraphrase of the true segment such as \emph{shoulder level} for \emph{shoulder height} replaces it with probability 0.5 while the negative keeps its wording, so new wording does not signal a negative~\cite{kamath2024hardpositive}. A clip without a partner receives up to 2 vocabulary swaps, negatives in which another word of the same scale replaces the stated value. Two further negatives join each with probability 0.5. One exchanges left and right in all texts and tracks, the other holds no text at all. Training draws clips with a rare high foot or wrist position up to 2.95 times more often. It takes about 21\,h on one H100 GPU.

\section{Experiments}
\label{sec:experiments}
\subsection{Setup}
\label{sec:setup}
\begin{figure*}[t]
\centering
\includegraphics[width=\textwidth]{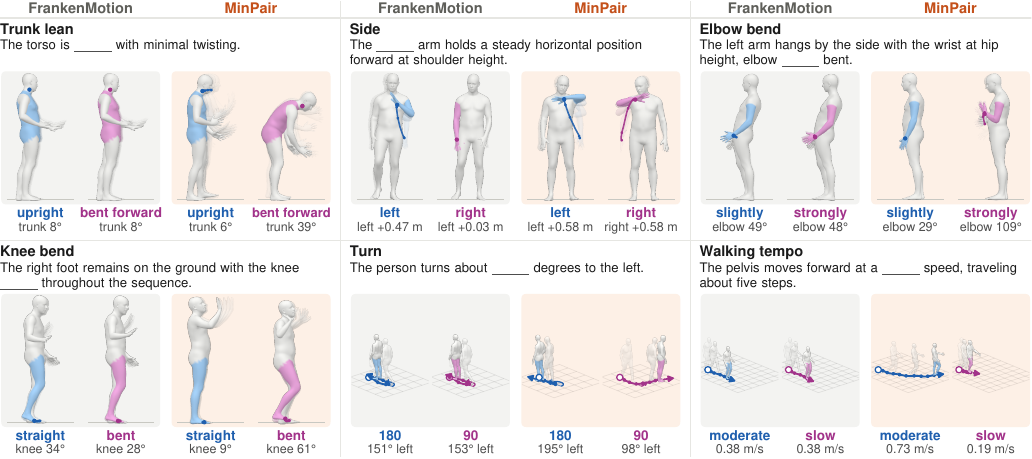}
\caption{\textbf{Caption edits.} Each block changes one attribute in the part captions of a held-out test clip (blue to magenta) under the same sampling noise. Elbow and knee are free-form edits, the others template edits. Turn and tempo show the pelvis path, the numbers the measured value. Over all test edits of these six kinds \method{} follows 83, 49, 72, 53, 48 and 43\%, FrankenMotion at most 3\%. Each block shows a clean edit with a large requested change whose motions come close to the requested values.}
\label{fig:edits}
\end{figure*}

\paragraph{Data.}
We use the AMASS~\cite{mahmood2019} subset of FrankenMotion~\cite{li2026b} with its split into 12,845 training clips of 30\,h in total, 1,651 validation and 1,601 test clips. Motions run at 20\,fps. We keep clips between 1 and 20\,s, which leaves 12,822 training and 1,596 test clips.

\paragraph{Baselines.}
We compare \method{} with four published generators. FrankenMotion~\cite{li2026b} is the part-track model whose input format \method{} shares. We use its released checkpoint, which is trained on the same AMASS split of 30\,h. Both part-track models use the same sampler without classifier-free guidance~\cite{ho2022cfg}. MARDM~\cite{meng2024mardm}, HY-Motion~1.0~\cite{wen2025hymotion} and Kimodo~\cite{rempe2026kimodo} read a single text. MARDM is a masked autoregressive diffusion model trained on HumanML3D~\cite{guo2022}. HY-Motion and Kimodo were trained on more than 3,000\,h and on 700\,h of motion. We run the official weights with the sampling settings of each release and resample the joints to 20\,fps. MARDM generates at most 9.8\,s and HY-Motion at most 12\,s. Kimodo generates every clip in one prompt at its length.

\paragraph{Inputs.}
Every model reads the input it was designed for. FrankenMotion and \method{} read \emph{part captions}, so a request is written into the timed segment of the relevant body part in the sentence style of \method{} captions. MARDM, HY-Motion and Kimodo read a single text, so they receive the whole request as \emph{one sentence}.

\paragraph{Target tasks.}
The first benchmark tests whether a model reaches a requested value in 22 tasks such as kick height, arm height, turn angle, walking distance, squat count, trunk lean and knee bend. Each task has three requested levels such as knee, waist or head height. Every level is written in both inputs and generated with 20 seeds shared across models. A script $m_q$ measures the quantity $q$ of the task, such as the peak foot height, in the motion $G(c_\ell,\xi)$ generated from the prompt $c_\ell$ of level $\ell$ with sampling noise $\xi$. A sample is a hit when this value falls into the target range $R_\ell$ of the requested level,
\begin{equation}
 \hit_A=\mathbf{1}\bigl[m_q(G(c_\ell,\xi))\in R_\ell\bigr].
 \label{eq:target}
\end{equation}
Attribute accuracy is the share of hits on the target tasks, averaged over the task and level cells of an input.

\paragraph{Caption edits.}
\begin{table*}[t]
\centering
\caption{\textbf{Comparison with published generators.} Each model reads its own input format. One-sentence inputs carry no side edits. Success and R@3 in \%, the best value of each column in bold.}
\label{tab:compare}
\fontsize{8.5}{10.3}\selectfont\renewcommand{\arraystretch}{1.0}\setlength{\tabcolsep}{4pt}
\begin{tabularx}{\textwidth}{@{}llrRRRRRRR@{}}
\toprule
Model & Input & Data (h) & \shortstack[r]{Target\\tasks $\uparrow$} & \shortstack[r]{Template\\edits $\uparrow$} & \shortstack[r]{Free-form\\edits $\uparrow$} & \shortstack[r]{All\\edits $\uparrow$} & M2M $\uparrow$ & R@3 $\uparrow$ & FID $\downarrow$ \\
\midrule
MARDM~\cite{meng2024mardm} & {\footnotesize\textcolor{tablesoft}{one sentence}} & $<$29 & 42.3 & 13.7 & 7.0 & 10.8 & 0.694 & 81.4 & 0.117 \\
HY-Motion~1.0~\cite{wen2025hymotion} & {\footnotesize\textcolor{tablesoft}{one sentence}} & $>$3,000 & 55.2 & 22.5 & 16.7 & 20.0 & 0.754 & \textbf{89.0} & 0.059 \\
Kimodo~\cite{rempe2026kimodo} & {\footnotesize\textcolor{tablesoft}{one sentence}} & 700 & 51.4 & 17.1 & 13.4 & 15.5 & 0.618 & 56.7 & 0.431 \\
FrankenMotion~\cite{li2026b} & {\footnotesize\textcolor{tablesoft}{part captions}} & 30 & 31.2 & 1.2 & 0.2 & 0.8 & 0.745 & 86.0 & 0.057 \\
\midrule
\method{} (ours) & {\footnotesize\textcolor{tablesoft}{part captions}} & 30 & \textbf{62.7} & \textbf{60.2} & \textbf{55.0} & \textbf{58.1} & \textbf{0.830} & 85.9 & \textbf{0.038} \\
\bottomrule
\end{tabularx}
\end{table*}

\begin{table*}[t]
\centering
\caption{\textbf{Effect of captions and pair loss.} Every row trains our denoiser, also the first, which uses the FrankenMotion captions and the FrankenMotion training procedure. Each row changes one setting of the row above it. The last row is \method{}. Success and R@3 in \%.}
\label{tab:main}\label{tab:bank}
\fontsize{8.5}{10.3}\selectfont\renewcommand{\arraystretch}{1.0}\setlength{\tabcolsep}{4pt}
\begin{tabularx}{\textwidth}{@{}lllRRRRRRR@{}}
\toprule
\multicolumn{3}{@{}l}{Our denoiser trained with} & Target & Template & Free-form & All & & & \\
\cmidrule(r){1-3}
Captions & Training & Pair loss & tasks $\uparrow$ & edits $\uparrow$ & edits $\uparrow$ & edits $\uparrow$ & M2M $\uparrow$ & R@3 $\uparrow$ & FID $\downarrow$ \\
\midrule
\textcolor{tablesoft}{FrankenMotion} & \textcolor{tablesoft}{FrankenMotion} & \textcolor{tablesoft}{--} & 27.5 & 1.4 & 0.1 & 0.9 & 0.729 & 81.6 & 0.089 \\
\method{} & \textcolor{tablesoft}{FrankenMotion} & \textcolor{tablesoft}{--} & 50.0 & 11.4 & 6.0 & 9.2 & 0.819 & 86.9 & 0.061 \\
\method{} & \method{} & \textcolor{tablesoft}{--} & 44.5 & 10.5 & 4.7 & 8.2 & \textbf{0.837} & \textbf{89.0} & \textbf{0.033} \\
\method{} & \method{} & template pairs & 60.7 & \textbf{61.4} & 27.6 & 47.7 & 0.834 & 88.0 & 0.034 \\
\method{} & \method{} & free-form pairs & 57.0 & 36.7 & \textbf{59.8} & 46.0 & 0.827 & 86.6 & 0.036 \\
\method{} & \method{} & both & \textbf{62.7} & 60.2 & 55.0 & \textbf{58.1} & 0.830 & 85.9 & 0.038 \\
\bottomrule
\end{tabularx}
\end{table*}

The second benchmark tests whether a single changed word changes the motion. A caption edit changes only text. It replaces the value of one attribute in the \method{} captions of a held-out test clip. The model then generates one motion from the unedited caption $c$ and one from the edited caption $c'$ with the same noise $\xi$, so a difference between the two motions comes from the text alone. The edits come from two sources. \emph{Template edits} are built by a script. They change arm height, leg height, tempo, turn, count or side in 3,292 edits of 1,574 clips and trunk lean, trunk twist, head speed or knee bend in 2,577 further edits by the rules of the template pairs. A side edit mirrors the left and right tracks. \emph{Free-form edits} come from the language model of the free-form pairs, which changes one attribute of 18 kinds in 4,007 edits. The script checks them like the free-form pairs. Let $\Delta_g=m_q(G(c',\xi))-m_q(G(c,\xi))$ be the generated change of the edited quantity $q$ and $\Delta_r$ the change observed between real clips with the two attribute values. An edit counts as \emph{followed} when
\begin{equation}
 \hit_B=\mathbf{1}\!\left[
  \operatorname{sign}(\Delta_r)\,\Delta_g\geq\tfrac12|\Delta_r|
 \right].
 \label{eq:response}
\end{equation}
The motion must move in the requested direction by at least half the real change. For the four posture attributes and the free-form edits $\Delta_r$ is the measured difference to one training partner, over segments for posture edits and over whole clips for free-form edits. Edit success is the share of followed edits of one source, where every edit counts once since some kinds hold only a few edits. \Cref{fig:edits} shows template edits of trunk lean, side, turn and walking tempo as well as free-form edits of elbow and knee bend.

\paragraph{Quality and alignment.}
We also report three metrics of the FrankenMotion evaluator~\cite{li2026b} on the 1,562 test clips with a global caption, averaged over three generation runs. M2M is the rescaled cosine $(1+\cos)/2$ between embeddings of the generated and the reference motion. R@3 is the top-3 text retrieval accuracy. FID is the Fr\'echet distance between embedding distributions. Each motion is generated from the complete caption of a real test clip and compared with that clip. Models trained on the FrankenMotion captions receive the FrankenMotion test captions and models trained on \method{} captions the held-back \method{} captions of the same clips, which the pipeline of \cref{sec:base} writes from the measurements of each test clip. The evaluator was trained on the FrankenMotion captions and scores every text metric against them. We therefore omit its motion-to-text similarity M2T and use M2M as our primary quality metric because it compares motion with motion. The three single-text models receive the global caption and the clip length.

\paragraph{Control runs.}
To separate the captions from the network, we also train our denoiser with the FrankenMotion training procedure and the denoising loss alone, once on the part captions of FrankenMotion and once on \method{} captions. Both runs share clips, split, settings and epochs and differ only in the captions.

\subsection{Comparison with published generators}
\label{sec:compare}\label{sec:routes}
\Cref{tab:compare} gives every model its own input. With part captions \method{} reaches 62.7\% attribute accuracy and follows 60.2\% of the template edits and 55.0\% of the free-form edits. This is more than any published generator reaches, although HY-Motion and Kimodo were trained on 23 to 99 times more motion data. When both part-track models instead read the one sentence of MARDM, HY-Motion and Kimodo in the global caption and in all eight tracks, FrankenMotion follows 10.0\% of the template edits and 5.0\% of the free-form edits. \method{} still follows 50.7\% and 50.0\%, more than any published generator. On the test clips of at most 9.8\,s, within the length limit of every model, \method{} follows 58.2\% of the edits against 0.9 to 19.7\% for the published generators.

\subsection{Effect of captions and pair loss}
\label{sec:results}
\begin{table*}[t]
\centering
\caption{\textbf{Ablation of the pair loss training.} Each row below \method{} changes one part of its training and trains from scratch. Success and R@3 in \%.}
\label{tab:ablation}
\fontsize{8.5}{10.3}\selectfont\renewcommand{\arraystretch}{1.0}\setlength{\tabcolsep}{4pt}
\begin{tabularx}{\textwidth}{@{}lRRRRRRR@{}}
\toprule
Model & \shortstack[r]{Target\\tasks $\uparrow$} & \shortstack[r]{Template\\edits $\uparrow$} & \shortstack[r]{Free-form\\edits $\uparrow$} & \shortstack[r]{All\\edits $\uparrow$} & M2M $\uparrow$ & R@3 $\uparrow$ & FID $\downarrow$ \\
\midrule
\method{} & \textbf{62.7} & \textbf{60.2} & 55.0 & \textbf{58.1} & 0.830 & 85.9 & 0.038 \\
\quad pair loss at high noise only & 57.6 & 49.4 & 44.3 & 47.3 & 0.829 & 87.0 & \textbf{0.036} \\
\quad pair loss at every step & 57.7 & 59.4 & \textbf{56.0} & 58.0 & \textbf{0.831} & \textbf{87.3} & \textbf{0.036} \\
\quad w/o vocabulary swaps & 58.3 & 58.5 & 53.9 & 56.6 & 0.827 & 85.9 & 0.039 \\
\quad w/o rare-position oversampling & 61.9 & 57.9 & 53.0 & 55.9 & 0.830 & 86.9 & 0.038 \\
\bottomrule
\end{tabularx}
\end{table*}

\paragraph{Captions and pair loss.}
\Cref{tab:main} trains the same denoiser in every row and changes one setting per row. Trained on the FrankenMotion captions, this denoiser is no better than FrankenMotion itself (\cref{tab:compare}). Its attribute accuracy is 27.5\% against 31.2\% and it follows 1.4\% of the template edits against 1.2\%. The network therefore does not explain the gains that follow. \method{} captions alone raise attribute accuracy to 50.0\% and template-edit success to 11.4\%. The training of \cref{sec:training} without the pair loss adds no control (44.5\% and 10.5\%). The pair loss with template pairs then raises template-edit success to 61.4\% and attribute accuracy to 60.7\%. \method{} captions thus teach the model which execution a complete caption describes. The pair loss makes it respond to a single changed word.

\paragraph{Pair sources.}
The last three rows of \cref{tab:main} differ only in their negative captions. Each source helps most on the edits it covers. Template pairs follow 61.4\% of the template edits but only 27.6\% of the free-form edits. Free-form pairs alone follow 59.8\% of the free-form edits but only 36.7\% of the template edits. \method{} combines both sources and follows 60.2\% of the template edits and 55.0\% of the free-form edits. Each source thus leads on its own edits, while \method{}, trained on both, leads in attribute accuracy with 62.7\% and on all edits together with 58.1\%.

\paragraph{Standard metrics.}
The standard metrics follow the captions but not the pair loss. Sequence M2M rises from 0.729 to 0.819 with \method{} captions and to 0.837 with the training of \cref{sec:training}. With the pair loss it is 0.830 although template-edit success rises from 10.5\% to 60.2\%. The reason is that these metrics test complete captions. A complete caption describes an execution redundantly. When we mask the arm-height, leg-height or tempo word of a held-out caption, a linear classifier fit on the training captions recovers it from the remaining words in 86\%, 73\% and 77\% of cases against 63\%, 60\% and 41\% for the most frequent value. A model can therefore produce the right execution from the context without using the attribute word. Motions generated from the unedited test captions indeed agree with the real clip equally often with and without the pair loss, in 87.8\% of cases over 8 measured quantities.

\subsection{Ablation of the pair loss training}
\label{sec:ablation}
\Cref{tab:ablation} changes one part of the pair loss training per row and trains from scratch. With the pair loss at high noise only, at the steps $t\geq50$, the model follows 49.4\% of the template edits and 44.3\% of the free-form edits instead of 60.2\% and 55.0\%. The less noisy steps therefore need the pair loss as well. With the pair loss at every step, the model follows 59.4\% of the template edits and 56.0\% of the free-form edits. Without vocabulary swaps template-edit success is 58.5\% and without rare-position oversampling, with all clips drawn equally often, 57.9\%.

\section{Conclusion}
\label{sec:conclusion}
We introduced \method{}, which builds the supervision for controllable text-to-motion generation from measurements of real motion clips. The same rule-based motion descriptors yield \method{} captions and the measured minimal pairs of a pair loss. With the same network, \method{} captions improve attribute accuracy, M2M, R@3 and FID over FrankenMotion captions, yet the denoiser still follows only 11\% of the template edits. The pair loss raises this share to 60\% while the three standard metrics do not improve further. Controlling a single attribute is thus a separate skill from fitting a complete caption. A model acquires it reliably only when one word alone separates two otherwise identical captions. With body-part captions \method{} also reaches more targets than state-of-the-art methods trained on far more motion, so targeted supervision can outweigh scale.

\paragraph{Impact.}
\method{} turns the words of a caption into controls. An animator can change how high, how fast or how far a movement goes by editing one word. The same control applies when a robot imitates the generated motion and its operator states the execution in words. The supervision needs neither new recordings nor new human annotation, only computation, so it scales with the size of the corpus. The pair loss adds one term to training and leaves inference unchanged, so existing diffusion generators can adopt it. The two benchmarks measure the control that standard metrics miss and apply to any text-to-motion model. Measured minimal pairs are not tied to motion. They can be built wherever the property behind a word can be measured.

\paragraph{Limitations.}
\method{} covers attributes that can be measured from joint positions and expressed by a word scale or a short phrase. Style, emotion and contact with objects lie outside this set. We trained the pair loss with one part-track denoiser on one corpus, so other generator families remain to be tested. Furthermore, applying the pair loss to all clips of the batch remains unexplored.

\paragraph{Future work.}
So far the measured words act through the part tracks. The next step carries them into the global caption so that a single sentence steers the execution as well. Further pairs can change when a movement happens or how several body parts move together, as long as each pair stays checkable against the measurements. The creation of the supervision can also become agentic, with language-model agents that find what the generator still misses and add the measurements and pairs for it.

\section*{Acknowledgments}
This research was funded by the European Regional Development Fund (ERDF) within the project `AIDA' (grant FEIH 2476184). Views and opinions expressed are those of the authors only and do not necessarily reflect those of the European Union. Neither the European Union nor the granting authority can be held responsible for them.

Writing and coding assistants (Claude by Anthropic, ChatGPT by OpenAI) helped to draft and revise this manuscript, write and run experiment code and prepare tables and figures. The authors verified every number, claim and reference and take full responsibility for the content.

{\small
\bibliographystyle{ieeenat_fullname}
\bibliography{references}
}
\end{document}